%% file: main.tex
\documentclass[sigconf,screen]{acmart}
\makeatletter
\global\@ACM@balancefalse
\makeatother
\usepackage[table]{xcolor}
\usepackage{arydshln}
\newcommand{\rev}[1]{#1}
\AtBeginDocument{%
  }

\setcopyright{acmlicensed}
\copyrightyear{2026}
\acmYear{2026}
\acmConference[MM '26]{Proceedings of the 34th ACM International Conference on Multimedia}{November 10--14, 2026}{Rio de Janeiro, Brazil}
\acmBooktitle{Proceedings of the 34th ACM International Conference on Multimedia (MM '26)}
\acmDOI{10.1145/XXXXXXX.XXXXXXX}
\acmISBN{979-8-4007-XXXX-X/2026/10}

\begin{document}

\title{EmoStyle: Affective Conditioning of Style-Specialist Experts for Emotional Image Generation}

\author{Dexiang Hong}
\email{hongdexiang@mail.ustc.edu.cn}
\affiliation{%
  \institution{University of Science and Technology of China}
  \country{HeFei, China}}

\author{Yijie Guo}
\email{guoyijie@ustc.edu}
\affiliation{%
  \institution{University of Science and Technology of China}
  \country{HeFei, China}}

\author{Weidong Chen}
\correspondingauthor
\email{chenweidong@ustc.edu.cn}
\affiliation{%
  \institution{University of Science and Technology of China}
  \country{HeFei, China}}

\author{Xinyan Liu}
\email{xinyliu@hit.edu.cn}
\affiliation{%
  \institution{Harbin Institute of Technology, WeiHai}
  \country{WeiHai, China}}

\author{Zixuan Zou}
\email{zouzixuan@hit.edu.cn}
\affiliation{%
  \institution{Harbin Institute of Technology, WeiHai}
  \country{WeiHai, China}}

\author{Zhendong Mao}
\email{zdmao@ustc.edu.cn}
\affiliation{%
  \institution{University of Science and Technology of China}
  \country{HeFei, China}}

\author{Yongdong Zhang}
\email{zhyd73@ustc.edu.cn}
\affiliation{%
  \institution{University of Science and Technology of China}
  \country{HeFei, China}}

\renewcommand{\shortauthors}{Hong et al.}

% \begin{abstract}
% In contrast to conventional text-to-image generation, emotion-aware artistic image generation requires the generated artwork to jointly satisfy semantic content, artistic style, and intended affect. This task is constrained by two main challenges. The first challenge is affective under-specification, where short prompts often leave valence-arousal tendency, dominant emotion, visual attributes, composition, and resolution ambiguous. The second challenge is style-emotion coupling, where emotional expression must be realized through style-sensitive visual factors such as color, brushwork, lighting, line, and composition without weakening content or category identity. To address these issues, we propose EmoStyle, a Z-Image-based framework that combines structured affective reasoning with bucket-specific style LoRA experts. \textcolor{red}{An LLM reasoner expands each prompt into a training-style affective plan, including semantic completion, visual attributes, emotional interpretation, valence-arousal cues, dominant emotion, and aspect ratio}. A lightweight affective encoder injects this plan into the denoising process through AdaLN-style modulation, while a deterministic style-bucket mapping selects the corresponding LoRA expert trained for that bucket. At inference time, a VLM-guided refinement loop further evaluates candidates and revises the generation state when residual failures remain. In Track 1 (Emotion-Aware Artistic Image Generation) of the AffectiveArt Challenge 2026, our USTC\_PI\_LAB\_TEAM submission secured first place.
% \end{abstract}

\begin{abstract}
Emotion-aware artistic image generation requires an image to match the input prompt, follow the specified artistic style, and convey the target emotion. In this challenge, the main difficulty is that the visual and affective attributes available in the training data are not explicitly provided at test time. Without these attributes, the generator has to decide not only what to depict, but also how the target emotion should be expressed through color, lighting, brushwork, composition, line, and layout. This creates a control gap between the available test prompt and the fine-grained conditions needed for emotion-aware artistic generation. To bridge this gap, we propose EmoStyle, a Z-Image-based framework that converts the input prompt into a structured generation state. An LLM reasoner first predicts affective cues (valence-arousal, dominant emotion, and therapeutic-effect labels) and an aspect-ratio decision. Instead of using these predictions only as additional prompt text, we encode the affective fields into an affective condition vector and inject it into the denoising blocks through AdaLN-style modulation. This allows the inferred control variables to directly guide the generation of intermediate features. Since emotional expression is also style-dependent, we further train a dedicated LoRA adapter for each artistic style bucket and select the corresponding expert during inference, enabling the same affective cues to be rendered with bucket-specific priors for color, texture, brushwork, and composition. Finally, a lightweight VLM-guided candidate selection step ranks the generated images based on prompt alignment, style consistency, emotional expression, and visual quality. In Track 1 of the AffectiveArt Challenge 2026, our USTC\_PI\_LAB\_TEAM submission achieved first place.
\end{abstract}




\maketitle

\section{Introduction}

Large-scale artificial intelligence models have rapidly advanced a wide
range of multimedia tasks, including efficient video understanding,
retrieval and recognition, vision-language reasoning, sentiment-aware
language generation, and controllable creative
design~\cite{meng2026audiovisual,qin2025querybased,chen2021cascade,chen2022multiattention,chen2023weakly,hong2021siamese,hong2022crnet,li2022self,hong2021generic,li2022end,li2022structured,gu2026structured,liu2025matching,wang2025combatting,ye2025improving,li2025rethinking,jin2024multigrained,jin2024d2net,liu2024bootstrapping,wang2023improving,li2024exploring,fu2024sentiment,zhou2025hierarchical,tian2023endtoend,han2023textstyle,zhao2023difference,lin2024prompting,wen2025information,wang2025feature}.
Emotion-aware artistic generation combines text-to-image synthesis,
computational aesthetics, and affective computing. Recent diffusion and
flow-based models have greatly improved prompt-conditioned image fidelity
and controllability~\cite{ho2020ddpm,dhariwal2021diffusion,rombach2022ldm,saharia2022imagen,esser2024sd3,zimage2025}.
\rev{Yet visual emotion datasets and artistic benchmarks show that perceived emotion depends not only on object semantics, but also on artistic style, color, lighting, brushwork, line, and composition~\cite{achlioptas2021artemis,affectiveart2026challenge,yang2023emoset}. Emotion-aware artistic generation must therefore preserve content, follow style, and express the target emotion through appropriate visual form.}

In the AffectiveArt Challenge, the training data contains rich visual and affective annotations, including visual attributes, valence-arousal values, dominant emotions, and therapeutic-effect labels, but such structured annotations are absent at test time. The model receives only the input prompt and target style bucket, and must infer the missing cues that determine how the target emotion should be expressed under the required artistic style. To address this issue, we propose EmoStyle, a Z-Image-based framework for emotion-aware artistic generation. Given a prompt and its target style bucket, EmoStyle first uses an LLM reasoner to construct a structured generation plan, including affective cues (valence-arousal, dominant emotion, and therapeutic-effect labels) and an aspect-ratio decision, following the broader idea that prompt optimization and expansion can improve controllable generation~\cite{hao2022promptist,wang2023reprompt,cao2023beautifulprompt,datta2023promptexpansion}. The affective fields are encoded into an affective condition vector and injected into the denoising network through AdaLN-style shift-scale modulation~\cite{peebles2023dit}. This allows emotion-related cues to guide intermediate generation features rather than being used only as extra prompt text. To preserve style consistency, we train one LoRA adapter for each style bucket and select the corresponding expert during inference. The selected LoRA provides bucket-specific priors for color, texture, brushwork, and composition, while the affective condition controls how the target emotion is expressed within those priors. Since a single sample may still fail to satisfy content, style, affect, and quality constraints simultaneously, we further introduce a lightweight VLM-guided test-time refinement step, drawing on recent VLM reasoning and test-time scaling studies~\cite{liu2023llava,bai2023qwenvl,snell2024testtime,brown2024largelanguagemonkeys}. It evaluates generated candidates according to prompt alignment, style consistency, affective consistency, and visual quality, and selects the best output without additional training.

\begin{table*}[!htbp]
\centering
\caption{Leaderboard of the ACM MM'26 AffectiveArt Challenge (Track 1). Our submission (Rank 1) is highlighted with a gray background.}
\label{tab:track1_results}
\resizebox{\textwidth}{!}{
\begin{tabular}{c l c c c c c c c}
\toprule
\textbf{Rank} &
\textbf{Participant} &
\textbf{Overall Score$\uparrow$} &
\textbf{FID$\downarrow$} &
\textbf{FID Score$\uparrow$} &
\textbf{AAS$\uparrow$} &
\textbf{Content$\uparrow$} &
\textbf{Style$\uparrow$} &
\textbf{Attribute$\uparrow$} \\
\midrule
\rowcolor{gray!20}
1 & USTC\_PI\_LAB\_TEAM & \textbf{0.80} & \textbf{66.12} & \textbf{0.60} & 0.99  & 0.99  & 0.99  & 1.00  \\
2 & EmoForge & 0.78  & 77.47  & 0.56  & 1.00  & 1.00  & 1.00  & 1.00  \\
3 & VIRlab & 0.78  & 75.56  & 0.57  & 0.99  & 0.98  & 0.99  & 0.99  \\
4 & edaich & 0.77  & 78.59  & 0.56  & 0.99  & 0.98  & 0.99 & 0.99  \\
5 & Latent Feeling & 0.76  & 87.51  & 0.53  & 1.00  & 0.99 & 1.00  & 1.00  \\
\bottomrule
\end{tabular}
}
\end{table*}

This work makes three main contributions:
\begin{itemize}
    \item \rev{We introduce an LLM-based affective planner that infers missing affective and layout cues from the input prompt and target style bucket.}

    \item \rev{We propose a framework on top of Z-Image that separates style selection from affective modulation, where artistic styles are modeled by bucket-specific LoRA experts and affective cues are injected into denoising blocks through AdaLN-style modulation.}

    \item \rev{We design a VLM-guided test-time refinement strategy that evaluates generated candidates, identifies residual failures in content, style, affect, or quality, and selects or regenerates the final output. With this framework, our submission ranked first in Track 1 of the AffectiveArt Challenge 2026.}
\end{itemize}

\input{related_work}

\input{method}

\section{Experiments}
\subsection{Challenge Setting and Evaluation Metrics}

We follow Track 1 of the AffectiveArt Challenge 2026, namely
Emotion-Aware Artistic Image Generation. Given a short caption that
describes the semantic content, artistic movement, and the desired
emotional state, participants must generate a single artistic image
that satisfies all three conditions. The track therefore evaluates
whether a generative model can preserve the requested visual content
while aligning with both the specified artistic style and the target
emotional state.

The official dataset for the challenge is EmoArt, a large-scale,
emotion-aware artwork dataset containing 132,664 painting images
from public-domain art sources. EmoArt covers 56 artistic styles and
provides fine-grained annotations for both visual appearance and
affective semantics, including content descriptions, five visual
attributes, valence-arousal values, dominant emotion labels, and
therapeutic-potential labels. The five visual attributes are
brushwork, composition, color, line, and light, which are closely
related to how emotion is expressed in artworks.

For local model selection and ablation, we construct a validation set
from the official training split, matching the per-style sample count
of the test set so that the validation distribution is consistent with
the final evaluation setting. For the Gongbi category, where the
training split contains only 32 images but the test set contains 178,
we use visually related China Image samples as substitutes in the
train and val sets.

The Challenge evaluation metrics include Frechet Inception Distance (FID)
and Attribute Alignment Score (AAS). FID measures the distance between the feature distributions of generated and real images~\cite{heusel2017ttur}, with lower values indicating better visual fidelity and distributional realism.

AAS evaluates how well the generated artwork aligns with the
target artistic and affective attributes. Following the challenge
protocol, the official AAS is computed using a MiniCPM-V-2.6
evaluator~\cite{yao2024minicpmv} fine-tuned on EmoArt annotations.
This evaluator predicts an attribute description for each generated
image and compares it with the ground-truth attribute text using
CLIP similarity~\cite{radford2021clip}. Since the official fine-tuned
evaluator is not available during local development, we use the
original MiniCPM-V-2.6 evaluator without additional fine-tuning for
validation experiments, while the final leaderboard score is obtained
from the official challenge evaluator. Higher AAS indicates better
conditional alignment.

\begin{table}[H]
\centering
\caption{Main Results and Cross-Backbone Comparison on Local Validation Set (P2A+R: Prompt-to-Affect + Refinement).}
\label{tab:main_cross_backbone}
\setlength{\dashlinedash}{2.0pt}
\setlength{\dashlinegap}{1.0pt}
\resizebox{\linewidth}{!}{
\begin{tabular}{llccccc}
\toprule
Model & Setting & FID$\downarrow$ & AAS$\uparrow$ & Content$\uparrow$ & Style$\uparrow$ & Attribute$\uparrow$ \\
\midrule
Z-Image & Base & 75.8357 & 0.7309 & 0.6711 & 0.6972 & 0.8243 \\
\arrayrulecolor{gray!45}
\hdashline
\arrayrulecolor{black}
GPT Image 2 & P2A + R & 75.0873 & 0.7616 & 0.6815 & \textbf{0.7549} & 0.8484 \\
Nano Banana 2 & P2A + R & 79.2556 & 0.7487 & 0.6450 & 0.7477 & 0.8535 \\
\arrayrulecolor{gray!45}
\hdashline
\arrayrulecolor{black}
Flux2-klein-base & Full EmoStyle & 69.4151 & 0.7435 & 0.6591 & 0.6814 & \textbf{0.8901} \\
Z-Image & Full EmoStyle & \textbf{58.6373} & \textbf{0.7864} & \textbf{0.7153} & 0.7548 & 0.8892 \\
\bottomrule
\end{tabular}
}
\end{table}

\subsection{Implementation Details}

We use Z-Image as the backbone. For each style bucket we train a
separate LoRA expert (rank 128, 3 epochs, learning rate
$1\times10^{-4}$) with the backbone frozen; in the subsequent
affective-control stage the backbone and style experts are kept fixed,
and only the lightweight affective encoder and modulation heads are
optimized to inject emotional cues into the denoising process. The LLM
reasoner in Prompt-to-Affect Planning is instantiated with
Gemini-3.5-Flash~\cite{geminiteam2026gemini}, which converts short
captions and target styles into structured affective plans, and the
VLM judge in VLM-Guided Candidate Refinement with Qwen3.5-397B-A17B,
which ranks the generated candidates.

\subsection{Main Results and Cross-Backbone Comparison}

We conduct experiments on the validation set using Z-Image as the
generation backbone and report FID, AAS, and the three AAS components. Table~\ref{tab:main_cross_backbone} reports the main results and the cross-backbone comparison. Compared with the Z-Image
base model, full EmoStyle reduces FID from 75.8357 to 58.6373 and
increases AAS from 0.7309 to 0.7864, indicating improved visual
realism and conditional alignment. The gains in content, style, and
attribute scores show that the proposed structured
conditioning helps the model better preserve the prompt semantics,
target style, and affective attributes.

In addition, to further evaluate the effectiveness of our method, we compare it with several open-source and closed-source generation models, including Flux2-klein-base, GPT Image 2, and Nano Banana 2. Because closed-source API models do not expose internal parameters
for bucket-specific LoRA adaptation and AdaLN-style modulation, we
apply only the model-agnostic components of our framework to these models.
Specifically, the structured affective plan is serialized as part of
the generation instruction, and the candidates are ranked by the same
VLM-guided refinement procedure, while the full EmoStyle architecture
is implemented on open-source backbones. Overall, Z-Image achieves the
best FID and AAS, Flux2-klein-base the highest attribute alignment, and
GPT Image 2 competitive style alignment.

\begin{table}[t]
\centering
\caption{Ablation study on the local validation set.}
\label{tab:ablation}
\resizebox{\linewidth}{!}{
\begin{tabular}{lccccc}
\toprule
Method & FID$\downarrow$ & AAS$\uparrow$ &
Content$\uparrow$ & Style$\uparrow$ & Attribute$\uparrow$ \\
\midrule
\textbf{EmoStyle(Ours)} & \textbf{58.6373} & 0.7864 & 0.7153 & 0.7548 & \textbf{0.8892} \\
\quad- w/o aspect-ratio prediction & 72.5912 & 0.7866 & 0.7052 & \textbf{0.7784} & 0.8761 \\
\quad- w/o prompt-to-affect planning & 68.4145 & \textbf{0.7957} & \textbf{0.7535} & 0.7715 & 0.8621 \\
\quad- w/o style-specific LoRA & 66.8951 & 0.7653 & 0.6851 & 0.7098 & 0.8741 \\
\quad- w/o candidate refinement & 60.7312 & 0.7751 & 0.6987 & 0.7451 & 0.8815 \\
\bottomrule
\end{tabular}
}
\end{table}

Figure~\ref{fig:qualitative_cross_backbone} presents qualitative results across different generation backbones. The examples are consistent with the quantitative results in Table~\ref{tab:main_cross_backbone}: Z-Image produces more balanced results in terms of content preservation, style consistency, and visual quality, other backbones show varying degrees of degradation in style or detail.

\subsection{Ablation Study}

Table~\ref{tab:ablation} evaluates the contribution of each component.
Removing aspect-ratio prediction causes the largest FID degradation,
from 58.6373 to 72.5912, indicating that layout planning is important
for matching the visual distribution of EmoArt. Removing
prompt-to-affect planning weakens attribute alignment from 0.8892
to 0.8621, suggesting that structured affective planning helps the
model realize fine-grained visual attributes. The higher content,
style, and AAS scores in this variant indicate that these automatic
metrics may favor direct prompt matching, while the full model
achieves better visual realism and attribute realization.

Removing the style-specific LoRA results in a clear drop in style
alignment, from 0.7548 to 0.7098, confirming the importance of
style-specialized adaptation. Removing candidate refinement also
degrades both FID and AAS, showing that VLM-guided candidate
selection improves final output quality. Overall, the full method
achieves the best FID and attribute alignment while maintaining
competitive content and style alignment, demonstrating a balanced
improvement in visual quality and conditional control.

\section{Conclusion}
We present EmoStyle, a Z-Image-based framework for emotion-aware
artistic image generation. Rather than relying solely on the input
prompt, EmoStyle constructs a structured affective generation state
that integrates prompt-to-affect reasoning, bucket-specific LoRA
selection, AdaLN-style affective modulation, and VLM-guided test-time
refinement, letting the generator jointly consider semantic content,
artistic style, visual attributes, emotion cues, and layout while
keeping the base diffusion backbone frozen. Experiments on EmoArt show
that the framework improves visual quality and conditional alignment
across different backbones, and ablations confirm the contributions of
layout planning, affective reasoning, style-specific adaptation, and
candidate refinement. In Track 1 of the AffectiveArt Challenge 2026,
our system achieved first place, demonstrating the effectiveness of
structured affective conditioning for controllable artistic image
generation.

\bibliographystyle{ACM-Reference-Format}
\bibliography{references}

\end{document}

%% file: related_work.tex
\section{Related Work}

\subsection{Visual Emotion Datasets}

Visual emotion analysis relies on datasets that connect visual
content with human affective responses. Early benchmarks such as
FI provide large-scale discrete emotion labels for image emotion
recognition~\cite{you2016building}, while EMOTIC annotates natural
scenes with both categorical emotions and continuous
valence-arousal-dominance dimensions~\cite{kosti2020emotic}.
For artworks, ArtEmis introduces emotion attributions and
natural-language explanations, enabling the study of why a painting
evokes a particular feeling~\cite{achlioptas2021artemis}. Recent
datasets further improve annotation richness. EmoSet provides
interpretable emotion-related attributes~\cite{yang2023emoset},
EmoVerse builds an MLLM-driven emotion representation dataset for
interpretable visual emotion analysis~\cite{guo2025emoverse}, and
EmoArt provides multidimensional annotations for
emotion-aware artistic generation, including painting styles,
visual attributes, and emotion
categories~\cite{zhang2025emoart}.

These resources have advanced emotion recognition, explanation,
and benchmarking, and recent emotional generation methods for
images and videos also show the importance of grounding abstract
affect in concrete visual content~\cite{yang2024emogen,yuan2025coemogen,ye2024dualpath,chen2026subjective}.
Beyond direct generation, a related line of work studies
structured affective reasoning and explanation, including
fine-grained emotion-cause extraction~\cite{chen2026towards,ye2025multiround},
consensus-prompted affective explanation
captioning~\cite{song2026bridging,zhang2026stimuli}, and
multimodal empathetic response generation~\cite{wang2026multiagent,chen2026facenet}.
However, their annotations do not directly specify how a compact challenge
caption should be mapped to controllable generation variables,
such as affective state, visual attributes, style prior, and aspect
ratio. EmoStyle uses structured reasoning and affective conditioning
to bridge this gap.

\subsection{LoRA Adaptation and Composition}

Parameter-efficient adaptation is widely used to specialize large text-to-image models without retraining all parameters. LoRA represents weight updates as low-rank matrices, reducing adaptation cost while preserving the base model prior~\cite{hu2021lora}. Related personalization methods, including Textual Inversion and DreamBooth, learn new visual concepts or subject identities from a small number of examples~\cite{gal2022textual,ruiz2023dreambooth}. Custom Diffusion and Mix-of-Show further extend this approach to multi-concept customization and shared-scene composition~\cite{kumari2022customdiffusion,gu2023mixofshow}. Style-oriented diffusion methods, such as InST, StyleAligned, and ArtAdapter, further show that compact conditions or adapters can transfer artistic appearance and preserve style consistency~\cite{zhang2022inst,hertz2023stylealigned,chen2023artadapter}. These works show that compact adaptation modules can efficiently encode new concepts and styles.

Recent studies also examine how multiple adapters interact. Multi-LoRA Composition proposes training-free strategies such as LoRA Switch and LoRA Composite~\cite{zhong2024multilora}, while LoRA-Composer shows that naive use of multi-LoRA can lead to concept confusion or concept vanishing~\cite{yang2024loracomposer}. LoraHub further explores dynamic LoRA composition across tasks~\cite{huang2023lorahub}. In creative graphic design, multi-conditional diffusion, layout-to-image generation, editable multi-layer poster generation, image-conditioned poster design, and generative parsing methods also highlight the need for explicit control over layout, elements, and editable visual structure~\cite{zhang2026creatidesign,zhang2025creatilayout,zhang2025creatiposter,hu2025dreamposter,chen2026creatiparser}. These methods mainly target subjects, identities, tasks, design elements, or general concepts, rather than affective artistic variables such as valence, arousal, dominant emotion, visual attributes, and style-sensitive resolution. In contrast, we train LoRA experts for style buckets and use affective conditioning to control emotional expression, keeping style selection and affective modulation separate.

\subsection{Expert-Based Style Specialization}

Mixture-of-Experts (MoE) models offer a general mechanism for conditional specialization. Sparsely-gated MoE activates a small subset of experts for each input~\cite{shazeer2017moe}, while Switch Transformers simplify the design by selecting the top-1 expert~\cite{fedus2021switch}. Beyond language models, V-MoE applies sparse expert routing to vision transformers~\cite{riquelme2021vmoe}, Expert Choice Routing improves load balancing by letting experts select tokens~\cite{zhou2022expertchoice}, and Graph-MoE combines memory-augmented routers with sparse experts for multivariate time-series anomaly detection~\cite{huang2025graphmoe}. Recent reviews summarize the broader design trade-offs of sparse experts, including capacity, routing stability, and transfer~\cite{fedus2022sparseexpertreview}. These works support the idea that different inputs can benefit from specialized parameter subsets.

However, learning a free-form router requires a clear routing objective and sufficient data for each expert. In our setting, EmoArt already provides style-bucket annotations, which serve as an explicit and reliable specialization signal. Recent adapter-based MoE methods, such as MoLE and MixLoRA, show that LoRA modules can be organized as reusable experts~\cite{wu2024mole,li2024mixlora}. Unlike these learned-routing methods, we train one style LoRA expert per bucket and use a deterministic bucket-to-expert mapping. This keeps style selection interpretable while leaving affective variation to the proposed conditioning module.

%% file: method.tex
\section{Method}

\subsection{Framework Overview}

We address Track 1 of the AffectiveArt Challenge 2026, where each
input caption specifies semantic content, artistic movement, and the
desired emotional state. EmoArt~\cite{zhang2025emoart} provides the
affective fields and style buckets used in our method. For the
$i$-th sample, we denote the input caption as $x_i$ and its target
style bucket as $y_i$.

Our goal is to generate artwork that is semantically faithful,
emotionally aligned, and consistent with the target artistic style.
Since test captions are compact, we formulate the task as structured
affective conditioning over bucket-specific style experts.

\begin{figure*}[t]
\centering
\includegraphics[width=\textwidth]{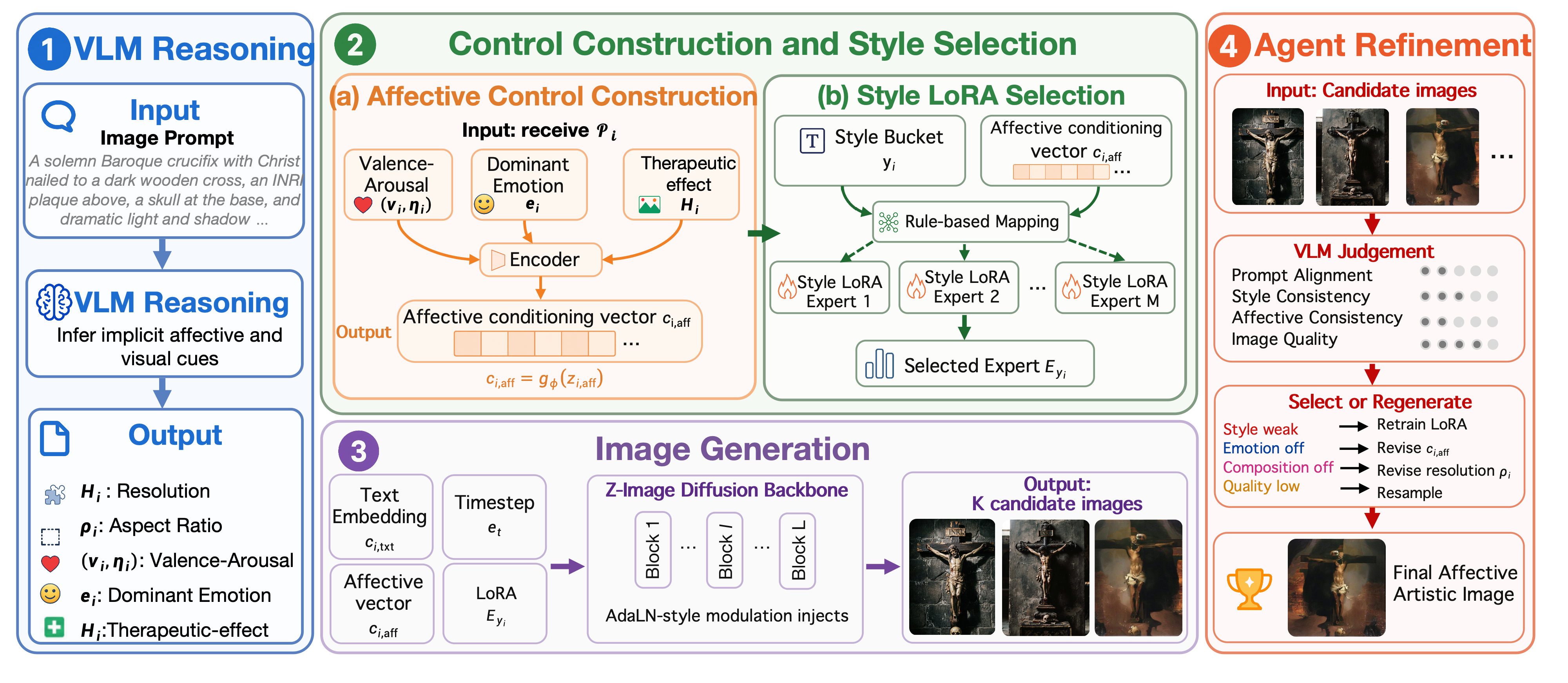}
\caption{Overview of EmoStyle. An LLM reasoner builds an affective plan, rule-based style selection and affective modulation condition the Z-Image generator, and a VLM judge refines the final candidate.}
\Description{Pipeline diagram with four stages: LLM reasoning, affective control construction and style-bucket LoRA selection, Z-Image generation, and VLM-guided candidate refinement.}
\label{fig:method_overview}
\end{figure*}

As shown in Fig.~\ref{fig:method_overview}, EmoStyle follows a
four-stage pipeline. First, an LLM reasoner expands the input caption
into an affective plan $\mathcal{P}_i$, including aspect ratio,
valence-arousal cues, dominant emotion, and therapeutic-effect labels.
Second, the affective fields are encoded as $\mathbf{c}_{i,\mathrm{aff}}$,
while a rule-based bucket mapping selects the style LoRA expert.
Third, Z-Image generates candidates from the text embedding, timestep
embedding, selected LoRA expert, and AdaLN-style affective modulation.
Finally, a VLM judge scores prompt alignment, style consistency,
affective consistency, and image quality, then selects the best
candidate or triggers regeneration when necessary.

\subsection{Prompt-to-Affect Planning}
\label{subsec:prompt-to-affect}

Although the challenge caption specifies semantic content, artistic
movement, and the desired emotional state, the structured affective
annotations available in EmoArt are not explicitly provided at test time.
Instead of rewriting the caption or generating free-form visual
attributes, we use an LLM reasoner only to infer compact affective and
layout variables that can be directly consumed by the generator.

For the $i$-th sample, the reasoner takes the input caption $x_i$, the
target style bucket $y_i$, and a fixed instruction template $\Pi$ as input:
\begin{equation}
(F_i,\rho_i) = R_{\psi}(x_i,y_i;\Pi),
\label{eq:prompt-to-affect}
\end{equation}
where $R_{\psi}$ denotes the LLM reasoner, $F_i$ is the predicted
affective summary, and $\rho_i$ is the predicted aspect ratio. Together
they constitute the affective plan $\mathcal{P}_i=(F_i,\rho_i)$.

The affective summary contains continuous valence-arousal scores and
discrete affective labels:
\begin{equation}
F_i=(\nu_i,\eta_i,e_i,H_i),
\label{eq:affective-summary}
\end{equation}
where $\nu_i,\eta_i\in[0,1]$ denote valence and arousal, $e_i$ is the
dominant emotion, and $H_i$ denotes the therapeutic-effect labels. The
dominant emotion and therapeutic-effect labels are selected from the
predefined EmoArt label vocabularies, while valence and arousal are
normalized continuous values.

The reasoner also predicts the aspect ratio $\rho_i$ from a predefined
candidate set. The predicted ratio determines the canvas orientation and
is converted into the final generation resolution. For landscape layouts,
we set the short edge based on the target style bucket and derive the long
edge from $\rho_i$; portrait layouts are handled symmetrically. The
resulting resolution is rounded to the nearest multiple of 16 before
generation.

The resulting plan provides two explicit control signals. The affective
summary $F_i$ is converted into the affective condition vector in
Sec.~\ref{subsec:affective-modulation} through valence-arousal coordinate embedding and label
embeddings, while the aspect-ratio decision $\rho_i$ determines the output
layout. The original caption $x_i$ is directly encoded by the frozen
Z-Image text encoder as the semantic condition. We do not use LLM-generated
caption rewriting or free-form visual attributes, which keeps the control
state compact and avoids relying on unstable natural-language fields at
test time.

\subsection{Bucket-Specific Style LoRA Experts}

EmoArt provides a target style bucket for each sample, which gives a
reliable supervision signal for style specialization. Instead of learning
an additional router, we train one LoRA expert for each style bucket on
top of the frozen Z-Image backbone. In our implementation, the bucket set
contains inkwash, ukiyoe, gongbi, renaissance, abstract, soviet realism,
baroque, and impressionism.

Let $p_i$ denote the text prompt of the $i$-th training sample, $I_i$
denote the corresponding artwork image, and $y_i \in \mathcal{Y}$ denote
its style bucket. For each bucket $y$, we define the bucket-specific
training subset as
\begin{equation}
\mathcal{D}_y = \{(p_i, I_i) \mid y_i = y\}.
\end{equation}
We train a LoRA expert $E_y$ with parameters $\theta_y$ on $\mathcal{D}_y$,
while keeping the Z-Image backbone $\Theta_Z$ frozen:
\begin{equation}
\theta_y^\star =
\arg\min_{\theta_y}
\mathbb{E}_{(p_i,I_i)\sim \mathcal{D}_y,\;t\sim \mathcal{U}(0,1)}
\left[
\mathcal{L}_{\mathrm{FM}}
\left(I_i,p_i,t;\Theta_Z,\theta_y\right)
\right],
\label{eq:bucket-lora-training}
\end{equation}
where $t$ is the flow-matching timestep and
$\mathcal{L}_{\mathrm{FM}}$ is the standard text-conditioned
flow-matching loss used by Z-Image. This bucket-wise training produces
a compact style prior for each artistic category while retaining the
general image-generation ability of the base model.

At inference time, style selection is deterministic:
\begin{equation}
E_i = E_{y_i}.
\label{eq:bucket-lora-selection}
\end{equation}
The selected expert is used throughout generation and remains fixed during
the subsequent affective-modulation stage. This design keeps style control
simple, avoids poorly supervised routing, and separates style
specialization from emotion-specific conditioning.

\subsection{Affective Condition Modulation}
\label{subsec:affective-modulation}

The structured plan in Sec.~\ref{subsec:prompt-to-affect} provides two types of affective cues:
continuous valence-arousal scores and discrete emotion labels. We do not
encode the natural-language visual attributes into the affective condition,
because these attributes are less stable at test time and may overlap with
the text prompt. Instead, we represent affect through a compact
label-coordinate embedding and inject it into the denoising blocks.

For sample $i$, let $\nu_i,\eta_i\in[0,1]$ denote valence and arousal,
$e_i$ denote the dominant emotion, and $H_i$ denote the set of
therapeutic-effect labels. We view valence and arousal as a
two-dimensional affective coordinate $\mathbf{a}_i = [\nu_i,\eta_i]^\top$.
To better represent this continuous affective space, we use a
sinusoidal coordinate embedding similar to positional encoding. Let
\begin{equation}
\mathcal{B}
=
\{(2^k,0),(0,2^k),(2^k,2^k),(2^k,-2^k)\}_{k=0}^{K-1}
\end{equation}
be a fixed set of frequency directions, where $K$ denotes the number of
frequency bands. The valence-arousal embedding is
defined as
\begin{equation}
\Phi_{\mathrm{VA}}(\mathbf{a}_i)
=
[
\mathbf{a}_i;
\{\sin(\pi \mathbf{b}^{\top}\mathbf{a}_i),
\cos(\pi \mathbf{b}^{\top}\mathbf{a}_i)\}_{\mathbf{b}\in\mathcal{B}}
].
\label{eq:va-fourier-embedding}
\end{equation}
This embedding preserves the continuous structure of the affective space
while allowing the model to distinguish different emotional regions and
intensity levels.

The dominant emotion is encoded with a learnable label embedding
$\mathbf{z}_{e,i}=E_e(e_i)$.
For the therapeutic-effect labels, which can be multi-label, we pool a
learnable per-label embedding $E_H$ over the active labels:
\begin{equation}
\mathbf{z}_{H,i}
=
\frac{1}{\max(1,|H_i|)}
\sum_{h\in H_i}E_H(h),
\label{eq:healing-label-embedding}
\end{equation}
where an empty label set is represented by a learnable null embedding.
The continuous affective coordinate is projected to the same dimension
by a learnable matrix $W_{\mathrm{VA}}$, i.e.
$\mathbf{z}_{\mathrm{VA},i}=W_{\mathrm{VA}}\Phi_{\mathrm{VA}}(\mathbf{a}_i)$.
We then build the affective feature by combining the continuous coordinate,
the discrete labels, and their interaction:
\begin{equation}
\mathbf{s}_{i,\mathrm{aff}}
=
[
\mathbf{z}_{\mathrm{VA},i};
\mathbf{z}_{e,i};
\mathbf{z}_{H,i};
\mathbf{z}_{\mathrm{VA},i}\odot \mathbf{z}_{e,i}
].
\label{eq:affective-feature-vector}
\end{equation}
The interaction term helps distinguish cases with similar
valence-arousal values but different dominant emotions. A lightweight
encoder maps this feature into the final affective condition:
\begin{equation}
\mathbf{c}_{i,\mathrm{aff}}
=
g_{\phi}(\mathbf{s}_{i,\mathrm{aff}}).
\label{eq:affective-condition}
\end{equation}

For each style bucket $y$, we pair its LoRA expert $E_y$ with a
bucket-specific residual modulation module $\mathcal{M}_y=\{M_{l,y}\}_l$.
The affective encoder $g_{\phi}$ is shared across buckets, while the
residual modulation heads are style-specific. This allows the model to
learn a common affective representation, while each style bucket learns
how to express the same affective state under its own visual prior.

We implement affective control as residual offsets to the native
scale-gate modulation of Z-Image, rather than inserting an additional
AdaLN layer. Let $h_{i,l}^{(t)}$ be the hidden state at denoising step $t$
and block $l$, and let $\mathbf{e}_t$ be the timestep embedding. The
frozen Z-Image block first produces its original modulation parameters:
\begin{equation}
(
\mathbf{s}_{l,\mathrm{msa}}^{Z,(t)},
\mathbf{g}_{l,\mathrm{msa}}^{Z,(t)},
\mathbf{s}_{l,\mathrm{mlp}}^{Z,(t)},
\mathbf{g}_{l,\mathrm{mlp}}^{Z,(t)}
)
=
A_l^Z(\mathbf{e}_t),
\label{eq:zimage-native-modulation}
\end{equation}
where $A_l^Z$ denotes the native modulation network of Z-Image. For the
active style bucket $y_i$, our affective head predicts residual offsets:
\begin{equation}
(
\Delta\mathbf{s}_{i,l,\mathrm{msa}}^{(t)},
\Delta\mathbf{g}_{i,l,\mathrm{msa}}^{(t)},
\Delta\mathbf{s}_{i,l,\mathrm{mlp}}^{(t)},
\Delta\mathbf{g}_{i,l,\mathrm{mlp}}^{(t)}
)
=
M_{l,y_i}(\mathbf{e}_t,\mathbf{c}_{i,\mathrm{aff}}).
\label{eq:affective-residual-modulation}
\end{equation}

The final scale and gate parameters are obtained by residual merging, for $\star\in\{\mathrm{msa},\mathrm{mlp}\}$:
\begin{equation}
\boldsymbol{\alpha}_{i,l,\star}^{(t)}
=
1+\mathbf{s}_{l,\star}^{Z,(t)}
+\Delta\mathbf{s}_{i,l,\star}^{(t)},
\quad
\boldsymbol{\tau}_{i,l,\star}^{(t)}
=
\tanh\!\left(
\mathbf{g}_{l,\star}^{Z,(t)}
+\Delta\mathbf{g}_{i,l,\star}^{(t)}
\right).
\label{eq:scale-gate}
\end{equation}

The selected style LoRA expert is applied inside the attention and MLP
layers of the block, where $\operatorname{RMSNorm}(\cdot)$ denotes
root-mean-square normalization. The resulting block computation is
\begin{equation}
\hat{h}_{i,l}^{(t)}
=
h_{i,l}^{(t)}
+
\boldsymbol{\tau}_{i,l,\mathrm{msa}}^{(t)}
\odot
\operatorname{RMSNorm}\!\left(
\operatorname{Attn}_{l,E_{y_i}}\!\left(
\operatorname{RMSNorm}(h_{i,l}^{(t)})
\odot
\boldsymbol{\alpha}_{i,l,\mathrm{msa}}^{(t)}
\right)\right),
\label{eq:affective-attn-residual}
\end{equation}
\begin{equation}
h_{i,l+1}^{(t)}
=
\hat{h}_{i,l}^{(t)}
+
\boldsymbol{\tau}_{i,l,\mathrm{mlp}}^{(t)}
\odot
\operatorname{RMSNorm}\!\left(
\operatorname{MLP}_{l,E_{y_i}}\!\left(
\operatorname{RMSNorm}(\hat{h}_{i,l}^{(t)})
\odot
\boldsymbol{\alpha}_{i,l,\mathrm{mlp}}^{(t)}
\right)\right).
\label{eq:affective-zimage-block}
\end{equation}

The final projection of each residual modulation head is zero-initialized,
so the residual offsets are initially zero. Therefore, at the beginning of
training, the block exactly reduces to the native Z-Image block equipped
with the selected style LoRA expert. During training, we keep the Z-Image
backbone and its native modulation network frozen, and jointly optimize
the shared affective encoder, the bucket-specific LoRA experts, and their
corresponding residual modulation modules using the standard Z-Image
flow-matching loss. For samples from bucket $y$, only $E_y$ and
$\mathcal{M}_y$ are updated, while $g_{\phi}$ receives gradients from all
buckets.

\begin{figure*}[!htbp]
\centering
\includegraphics[width=0.9\linewidth]{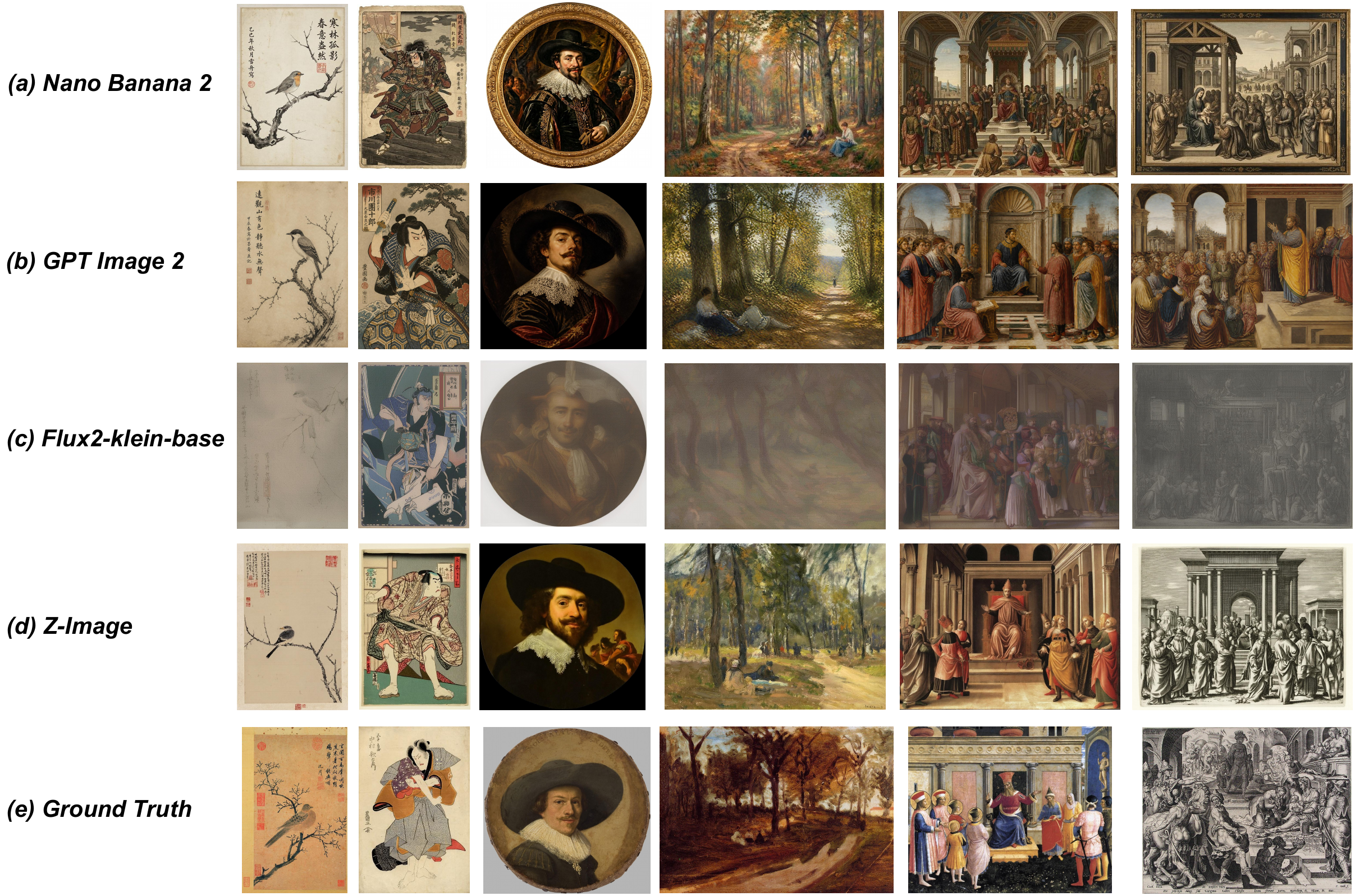}
\caption{Qualitative comparison of optimized results across different generation backbones. Each row corresponds to one model after applying our affective planning and refinement strategy, and each column represents one test sample with the corresponding ground-truth artwork shown in the last row.}
\Description{Qualitative comparison grid showing generated artworks from different backbones and the corresponding ground-truth artwork.}
\label{fig:qualitative_cross_backbone}
\end{figure*}

\subsection{VLM-Guided Candidate Refinement}

Although the planned prompt and affective conditions provide an initial generation state, a single stochastic sample may still
fail in content, style, affective expression, or image quality.
Therefore, during inference, we generate multiple candidates for
each prompt with different random seeds and rank them using a
VLM judge. Let $\mathcal{C}_i$ denote the generated candidate set for sample $i$. Given the input caption, target style, and affective plan,
the judge scores each candidate across four aspects: prompt
consistency, style consistency, affective-plan consistency, and
no-reference visual quality:
\begin{equation}
J(I)=\lambda_p J_p(I)+\lambda_s J_s(I)+
\lambda_a J_a(I)+\lambda_q J_q(I).
\end{equation}
Here, $\lambda_p$, $\lambda_s$, $\lambda_a$, and $\lambda_q$ are nonnegative weights; $J_p$ measures consistency with the input caption, $J_s$
measures consistency with the target style, $J_a$ measures
consistency with the affective plan, and $J_q$ measures
no-reference visual quality. The candidate with the highest score is
selected as the final output:
\begin{equation}
I_i^\ast=\arg\max_{I\in \mathcal{C}_i} J(I).
\end{equation}

This refinement strategy improves the final output without additional
training. It is especially useful for correcting occasional failures
caused by weak style expression, insufficient affective-plan
consistency, or poor visual quality.